\documentclass{article}

\PassOptionsToPackage{numbers}{natbib}

\usepackage[final]{drfair_edsc}

\usepackage[utf8]{inputenc} % allow utf-8 input
\usepackage[T1]{fontenc}    % use 8-bit T1 fonts
\usepackage[colorlinks=true]{hyperref}       % hyperlinks
\usepackage{url}            % simple URL typesetting
\usepackage{booktabs}       % professional-quality tables
\usepackage{amsmath}        % extended math environments
\usepackage{amsthm}         % extended math theorems 
\usepackage{amsfonts}       % blackboard math symbols
\usepackage{nicefrac}       % compact symbols for 1/2, etc.
\usepackage{microtype}      % microtypography
\usepackage{booktabs}       % nice tables
\usepackage{graphicx, caption, subcaption}  % sub figures

\newcommand{\A}         {\ensuremath{A}}
\newcommand{\F}         {\ensuremath{f}}
\newcommand{\Y}         {\ensuremath{Y}}
\newcommand{\Sc}        {\ensuremath{S}}
\newcommand{\X}         {\ensuremath{X}}
\newcommand{\x}         {\ensuremath{x}}
\newcommand{\Ai}        {\ensuremath{a}}
\newcommand{\Yi}        {\ensuremath{y}}
\newcommand{\Sci}       {\ensuremath{s}}
\newcommand{\N}         {\ensuremath{n}}

\newcommand{\As}        {\ensuremath{\mathcal{A}}}

\newcommand{\Ys}        {\ensuremath{\mathcal{Y}}}
\newcommand{\Xs}        {\ensuremath{\mathcal{X}}}
\newcommand{\Scs}       {\ensuremath{\mathcal{S}}}

\newcommand{\fn}[1]     {\ensuremath{\F\!\left({#1}\right)}}
\newcommand{\pr}[1]     {\ensuremath{\mathrm{P}\!\left({#1}\right)}}
\newcommand{\pc}[2]     {\ensuremath{\mathrm{P}\!\left({#1}|{#2}\right)}}
\newcommand{\Ex}[2]     {\ensuremath{\mathbb{E}_{#1}\!\left[{#2}\right]}}
\newcommand{\MI}[2]     {\ensuremath{\mathrm{I}\!\left[{#1};{#2}\right]}}
\newcommand{\NMI}[2]    {\ensuremath{\tilde{\mathrm{I}}\!\left[{#1};{#2}\right]}}
\newcommand{\CMI}[3]    {\ensuremath{\mathrm{I}\!\left[{#1};{#2}|{#3}\right]}}
\newcommand{\NCMI}[3]   {\ensuremath{\tilde{\mathrm{I}}\!\left[{#1};{#2}|{#3}\right]}}
\newcommand{\Ent}[1]    {\ensuremath{\mathrm{H}\!\left[{#1}\right]}}
\newcommand{\CEnt}[2]   {\ensuremath{\mathrm{H}\!\left[{#1}|{#2}\right]}}
\newcommand{\ratio}[1]  {\ensuremath{r_\text{#1}}}
\newcommand{\aratio}[1] {\ensuremath{\hat{r}_\text{#1}}}
\newcommand{\ami}[1]    {\ensuremath{\hat{I}_\text{#1}}}
\newcommand{\clf}[2]    {\ensuremath{\rho\!\left({#1} | {#2}\right)}}
\newcommand{\intd}[1]   {\ensuremath{\mathrm{d}{#1}}}
\newcommand{\Norm}[2]   {\ensuremath{\mathcal{N}\!\left({#1}, {#2}\right)}}

\newtheorem*{statement}{Problem Statement}

\title{Fairness Measures for Regression via Probabilistic Classification}

\author{%
  Daniel Steinberg$^*$, Alistair Reid$^\dagger$, Simon O'Callaghan$^\dagger$\\
  Gradient Institute\\
  $^*$Canberra \& $^\dagger$Sydney, Australia \\
  \texttt{\{dan, alistair, simon\}@gradientinstitute.org} \\
}

\begin{document}

\maketitle

\begin{abstract}
  Algorithmic fairness involves expressing notions such as equity, or
  reasonable treatment, as quantifiable measures that a machine learning
  algorithm can optimise.
  Most work in the literature to date has focused on classification problems
  where the prediction is categorical, such as accepting or rejecting a loan
  application.
  This is in part because classification fairness measures are easily computed
  by comparing the rates of outcomes, leading to behaviours such as ensuring
  that the same fraction of eligible men are selected as eligible women.
  But such measures are computationally difficult to generalise to the
  continuous regression setting for problems such as pricing, or allocating
  payments.
  The difficulty arises from estimating conditional densities (such as the
  probability density that a system will over-charge by a certain amount).
  For the regression setting we introduce tractable approximations of the
  independence, separation and sufficiency criteria by observing that they
  factorise as ratios of different conditional probabilities of the protected
  attributes.
  % Note - changed this from density ratio to conditional probabilities
  We introduce and train machine learning classifiers, distinct from the
  predictor, as a mechanism to estimate these probabilities from the data.
  This naturally leads to model agnostic, tractable approximations of the
  criteria, which we explore experimentally.
\end{abstract}

\section{Introduction}

A machine learning algorithm trained with a standard objective such as maximum
accuracy can produce behaviours that people, including those affected by its
decisions, consider unfair.
Algorithmic group fairness seeks to address this problem by defining protected
attributes such as age, race or gender, and introducing mathematical fairness
measures to evaluate and compare between the protected groups.
Many fairness measures have been proposed in the literature, but most can be
posed as specialisations of three general criteria, \emph{independence},
\emph{separation} and \emph{sufficiency}, that can be cleanly and intuitively
defined in terms of statistical independence~\cite{barocas_2019}.

%Independence requires the model predictions to be \emph{unconditionally 
%statistically} independent of the protected attribute.
%For example, you might ensure the acceptance rate of job applicants to be the
%same for males and females.
%This doesn't consider eligibility, making it both robust to historical
%discrimination in the recorded targets, but also blind to how prediction errors
%are distributed.
%%
%Separation requires model predictions to be independent of the
%protected attribute \emph{given the target variable}.
%This allows a relationship to exist between the score and the sensitive
%attribute to the extent justified by the target.
%A consequence is that two equally eligible people should receive the same
%score regardless of gender.
%%
%Sufficiency then requires the target to be independent of the protected
%attribute given the model's score.
%While this resembles separation in its phrasing, it states that two people
%receiving an equally high score should be equally likely to be eligible
%regardless of gender.

To date, the algorithmic fairness literature has mainly focused on
classification problems, where the decisions of a system
are binary or categorical, such as predicting who should be released on
bail~\cite{propublica_2016}, or who should be audited for a potentially
fraudulent tax return.
In this setting, fairness measures are straightforward to compute based on
% group-wise confusion matrices (by directly counting 
the fraction of each group that were correctly or incorrectly selected or
omitted.

However, there are many impactful regression problems
% where we have continuous
%or non-mutually exclusive outcomes we wish to predict, such as deciding how
%target 
such as how much to lend someone on a credit card or home loan,
or how much to charge for an insurance premium.
In the continuous setting, the aforementioned fairness criteria are generally
intractable to evaluate.
% TODO: below sentence needs citation!
Various tractable fairness measures have been proposed, but we believe that
many are either
(a) not as intuitive to reason about and thus apply as independence,
separation and sufficiency, or 
(b) are simplifications of these that fail to fully capture the important
properties of the original criteria.

\subsection{Contributions}

In this paper we introduce techniques to tractably numerically approximate the
\emph{independence}, \emph{separation} and \emph{sufficiency} group fairness
criteria in the general regression setting, in a way that is \emph{agnostic} to
the regression algorithms used.
% The approximation does not require simplification of the criteria \emph{beyond}
% the assumption that the groups, or sensitive attributes, are categorical in
% nature.
In particular, our contributions are,
\begin{itemize}
  \item 
    We approximate fairness criteria for a regression group fairness setting
    by factorising them as conditional
    probabilities of the (categorical) protected attribute,
    and apply estimation techniques based on probabilistic
    classification~\cite{qin_1998} and empirical integration to estimate
    their values.
    We explore two measures:
    (1) A density ratio measure that is readily applicable to binary sensitive
    attribute settings.
    (2) A more general (conditional) mutual information approximation that
    is applicable to categorical sensitive attributes and is normalised to a
    range of $[0, 1]$, with 0 being perfectly fair and 1 being maximally unfair
    according to the relevant criterion.
  \item Using these ratio estimation techniques to provide insight into how
    group fairness criteria operate in a general regression setting.
    We do this in a synthetic scenario, and benchmark on an existing fair
    regression algorithm to asses how it addresses the three group fairness
    criteria.
    % ; this has not been demonstrated previously in the literature.
\end{itemize}

Due to the general nature of these fairness criteria, we foresee them being
useful in many scenarios.
Inprocessing model predictions based on the criteria is left to future work.
If the predictor may not access the sensitive attributes,
preprocessing approaches may then be used to learn a fair representation, 
potentially incurring additional algorithmic performance and fairness
costs~\cite{mcnamara_2019}.
They are also well suited to the regulation and auditing of existing systems,
as they only require access to the prediction, outcome and sensitive
attributes, while the predictor itself can be a black box system, or even
incorporate \emph{manual} decisions.

\subsection{Related work}

Predictive fairness measures date back more than fifty years in the context
of testing and hiring, as summarised by~\citet{hutchinson_2019}.
Of particular relevance to our work are the measures of regression fairness
based on (partial) correlations between the predicted score, the target
variable and the demographic group of applicants, as derived
in~\cite{cleary_1968, darlington_1971}.
These measures can be interpreted as relaxations of the independence,
separation and sufficiency criteria under 
% Gaussianity assumptions.
the assumption that the outcome, prediction and group are jointly Gaussian distributed.
Propensity score stratification techniques have also been used to 
% applied to linear regression models, 
decrease the separation of the protected class for linear
regressors in~\cite{calders_2013}.
\citet{berk_2017} propose fairness objectives that address separation using
convex regularisers.
One of these regularisers (`group') permits cancelling of model errors within a
group, addressing an imbalance in expectation, while the other (`individual')
does not.
\citet{fitzsimons_2019} have developed an inprocessing technique for kernel
machines and decision tree regressors with a focus on satisfying (a relaxation of) the independence criterion.
% such that they also can be made to satisfy the independence criterion in expectation.
% TODO: mention HSIC?
Recently \citet{williamson_2019} proposed that risk measures from mathematical
finance generalise to fairness measures by imposing that the distribution of
losses across subgroups are commensurate.
% This general notion of group fairness examines the \emph{distribution} of
% a base loss function across groups.
% This constru a general fairness framework, and using 
%to evaluate the distributional properties of losses results in a
Of particular interest, the conditional value at risk measure leads to a convex
inprocessing technique for both regression and classification problems.

Mutual information (MI) has also previously been considered in the context of fair
supervised learning.
\citet{kamishima_2012} use normalised MI to asses fairness in
their \emph{normalised prejudice index} (NPI).
Their focus is on binary classification with binary sensitive attributes, and
the NPI is based on the independence fairness criterion.
In such a setting mutual information is readily computable empirically from
confusion matrices.
This work is generalised in~\cite{fukuchi_2015} for use in regression models by
using a \emph{neutrality} measure, which the authors show is equivalent to the
independence criterion.
They then use this neutrality measure to create inprocessing techniques for
linear and logistic regression algorithms.
Similarly,~\citet{ghassami_2018} take an information theoretic approach to
creating an optimisation algorithm that returns a predictor score that is fair
(up to some $\epsilon$) with respect to the separation criterion.
These works are mainly concerned with computing MI (or some
approximation) in the optimisation objective of \emph{particular algorithms}.
%insofar as it can be used for inprocessing techniques for particular algorithms.
Our main focus in this paper is to compute (conditional) mutual information in
a way that is \emph{agnostic} to the prediction algorithm used.

\section{Problem formulation}

We consider a predictor with target set, $\Ys$, feature
set, $\Xs$, and a sensitive attribute set, $\As$.
Our data is composed of random variables $\Y$, $\A$ and $\X$ drawn from some
distribution over $\Ys \times \Xs$
($\As$ is a subset of $\Xs$).
Here $\F$ is the prediction function % (regressor) of
that maps $\F : \Xs \to \Scs$, and results in predictions, or scores,
$\Sc = \fn{\X}$, where usually $\Scs = \Ys$.
For the purposes of this paper we will treat $\Sc$ as a random variable drawn
from a distribution over $\Scs$.

% We limit ourselves to approximating three general group fairness criterion,
We are specifically interested in approximating \emph{independence}, 
\emph{separation} and \emph{sufficiency}, a selection that subsumes
many other measures as relaxations or special cases~\cite{barocas_2019}. These
can be defined in terms of independence or conditional independence statements,
\begin{align}
  \text{Independence}&: \quad \Sc \bot \A \hspace{4em} \text{Equivalently},
     &&\pr{\Sc, \A} = \pr{\Sc}\pr{\A}, \label{eq:ind} \\
  \text{Separation}&: \quad \Sc \bot \A~|~\Y 
     &&\pc{\Sc, \A}{\Y} = \pc{\Sc}{\Y}\pc{\A}{\Y}, \label{eq:sep} \\
  \text{Sufficiency}&: \quad \Y \bot \A~|~\Sc
     &&\pc{\Y, \A}{\Sc} = \pc{\Y}{\Sc}\pc{\A}{\Sc}. \label{eq:suf}
\end{align}
Rather than constraining the predictor such that these conditions be exactly
satisfied, we would usually define a continuous measure of the degree to which
they are satisfied. This admits a balancing of priorities between these and other
measures such as accuracy.

In the \emph{classification} setting where $\Y$, $\Sc$ and $\A$ are all
categorical, these conditional probabilities can be empirically estimated from
confusion matrices between $\Y$ and $\Sc$ for each protected subgroup
($\A$)~\cite{barocas_2019}.
%%
%However, many real-world decision systems that affect people are actually
%regressors, with continuous targets. 
%For example, insurance pricing and credit risk
%scoring are concerned with predicting a continuous (often bounded) outcome,
%$\Sc$. Computing these group fairness measures in this case is difficult,
%even if we \emph{still assume} binary or categorical sensitive
%attributes. The difficulty arises from the need to estimate probability
%densities. For example, consider the separation criterion~\eqref{eq:sep},
%\begin{align*}
%  \pc{\Sc}{\A, \Y} &= \pc{\Sc}{\Y} \quad
%    \text{where we have divided~\eqref{eq:sep} by } \pc{\A}{\Y},\\
%  \pc{\Sc}{\A=1, \Y} &= \pc{\Sc}{\A=0, \Y} = \pc{\Sc}{\Y}
%    \quad \text{for} \quad \As = \{0, 1\}.
%\end{align*}
% In the classification setting, these conditional probabilities are a scalar
% quantity that is easy to estimate from frequency of occurrence.
In a regression setting, however, they become continuous density functions,
and we are required to either assume a parametric form for them, or numerically
estimate the densities using methods such as kernel density
estimation~\cite{hastie2001}.
% Al says: i'm not sure the sentence below is relevant?
% Unfortunately these methods become intractable very quickly once we move past a
% few dimensions with only a limited number of samples~\cite{hastie2001}.
% Short version:

Others have simplified these criteria using conditional expectation instead
of conditional probability~\cite{calders_2013, zafar_2015, fitzsimons_2019}.
But these approaches can fail to capture effects, such as groups with
different score variances, that can still lead to harm.
% Long version:
% Some approaches in the literature apply numerical techniques to evaluate
% expectations, $\Ex{}{\Sc | \A=1, \Y}$ $= \Ex{}{\Sc | \A=0, \Y}$, and examine
% measures of deviation between them. This is the basis of the approaches taken
% for computing the approximate fairness criterion in~\cite{calders_2013,
% zafar_2015, fitzsimons_2019}\footnote{Although some of these works go on to
% approximate fairness criteria from higher order moments or even covariances}.
% The problem with this approach is that 
% it fails to look at the score variation.
% For example the system may be unreliable with high estimation variance for one
% group, and reliable with a low variance for another.
% So long as the variance is unbiased, system may appear to
% have ideal separation fairness from the expectation perspective.
%
Our primary aim is to create techniques that tractably approximate the three
criteria of group fairness aforementioned in a way that is agnostic to the
predictor used to generate the score, $\Sc$. More formally,

\begin{statement}
  Derive methods that can approximate the fairness criteria,
  \emph{independence}, \emph{separation} and \emph{sufficiency}, for $\Ys
  \subseteq \mathbb{R}^d$ and $\Scs \subseteq \mathbb{R}^d$, in the case of
  \emph{binary} or \emph{categorical} sensitive attributes, $\As = \{0, 1\}$ or
  $\{0, \ldots, K\}$.
\end{statement}

\section{Direct Density Ratio Estimation}%
\label{sec:ratio}

By dividing both sides of~\eqref{eq:ind}-\eqref{eq:suf} by the distributions
$\pc{\A}{\cdot}$ and firstly considering $\As = \{0, 1\}$, we can re-express the
fairness criteria as density ratios,
\begin{align}
  \ratio{ind} = \frac{\pc{\Sc}{\A=1}}{\pc{\Sc}{\A=0}}, \quad
  \ratio{sep} = \frac{\pc{\Sc}{\A=1, \Y}}{\pc{\Sc}{\A=0, \Y}}, \quad
  \ratio{suf} = \frac{\pc{\Y}{\A=1, \Sc}}{\pc{\Y}{\A=0, \Sc}}.
  \label{eq:ratios}
\end{align}
Computing these ratios empirically is a common way to measure the fairness
criteria in classification~\cite{barocas_2019}, where perfect fairness would
correspond to a constant ratio of $1$.
% AL's note: should we introduce these as expectations from the beginning
% what do we loose by taking the expectation of a ratio? 
% doesn't it mean that a blip of much higher density can cancel out 
% a large amount of lower density
% Dan's response: We could do, though this is implying the population density
% ratio, at least in a frequentist sense... perhaps this needs to all be
% formalised a little more?
In a regression setting it is much easier to estimate the ratios
in~\eqref{eq:ratios} directly using \emph{density ratio estimation}
methods~\cite{sugiyama_2010, sugiyama_machine_2012},
as opposed to independently estimating the probability densities
in~\eqref{eq:ratios} then calculating the ratio.  
Density ratio estimation has wide applicability, for example it has been used to
detect and correct covariate shift~\cite{bickel_2009}, two sample
testing~\cite{qin_1998}, and compute divergence measures such as Kullback Leibler
divergence as well as mutual information measures~\cite{sugiyama_2010,
sugiyama_density_2012}.

We will now derive a density ratio estimator for $\ratio{ind}$
in~\eqref{eq:ratios} with binary sensitive attributes. First, begin by
recalling Bayes' Theorem, $\pc{\Sc}{\A} =
\pc{\A}{\Sc}\pr{\Sc}\!/\pr{\A}$, and then substituting into
$\ratio{ind}$,
\begin{align}
  \ratio{ind} = \frac{\pc{\A=1}{\Sc}\pr{\A=0}}{\pc{\A=0}{\Sc}\pr{\A=1}}.
  \label{eq:r_ind}
\end{align}
We could have also arrived at this directly from~\eqref{eq:ind} by dividing
both sides by $\pr{\Sc}$ such that $\pc{\A}{\Sc} = \pr{\A}$.
Pragmatically it would be more useful to know this ratio in expectation, or
even approximately as an empirical average over the data we have,
\begin{align}
  \Ex{\Sc}{\ratio{ind}} \approx \frac{\N_0}{\N_1} \cdot \frac{1}{\N}
  \sum_{i=1}^\N \frac{\pc{\A=1}{\Sc=\Sci_i}}{\pc{\A=0}{\Sc=\Sci_i}}.
  \label{eq:er_ind}
\end{align}
Here we denote instances of random variables by their lower case. $\N$ is the
total number of instances in our dataset and we have approximated
$\pr{\A=0}\!/\pr{\A=1}$ as the ratio of the number of instances in each
category of the sensitive attribute, $\N_0 / \N_1$.
A well known technique is to approximate density ratios directly with the
output of a probabilistic classifier~\cite{qin_1998, bickel_2009,
sugiyama_2010}. In this case we can use,
\begin{align}
  \pc{\A = a}{\Sc=\Sci} &\approx \clf{a}{\Sci},
\end{align}
where $\clf{\Ai}{\Sci}$ is a prediction of the probability that class $\A=a$
made by introducing and training a machine learning classifier (distinct from
the machine learning model $f$, which predicts the target).
Now we can complete the approximation~\eqref{eq:er_ind} numerically,
\begin{align}
  \Ex{\Sc}{\ratio{ind}} \approx \aratio{ind}
    = \frac{\N_0}{\N_1 \N} \sum_{i=1}^\N
    \frac{\clf{1}{\Sci_i}}{1 - \clf{1}{\Sci_i}}.
    \label{eq:a_ind}
\end{align}
Noting that in the binary classification case, $\clf{0}{\cdot} = 1 -
\clf{1}{\cdot}$. We can use similar density ratio estimates
to approximate $\ratio{sep}$ and
$\ratio{suf}$; again making use of Bayes' rule, $\pc{\Sc}{\A, \Y} =
\pc{\A}{\Y, \Sc}\pc{\Sc}{\Y} / \pc{\A}{\Y}$ and $\pc{\Y}{\A, \Sc} =
\pc{\A}{\Y, \Sc}\pc{\Y}{\Sc} / \pc{\A}{\Sc}$, we can rewrite,
\begin{align}
  \ratio{sep} = \frac{\pc{\A=1}{\Y, \Sc} \pc{\A=0}{\Y}}
    {\pc{\A=0}{\Y, \Sc} \pc{\A=1}{\Y}}, \qquad
  \ratio{suf} = \frac{\pc{\A=1}{\Y, \Sc} \pc{\A=0}{\Sc}}
    {\pc{\A=0}{\Y, \Sc} \pc{\A=1}{\Sc}}.
\end{align}
Now we make use of two more probabilistic classifiers $\clf{1}{\Yi, \Sci}$ and
$\clf{1}{\Yi}$ to approximate $\Ex{\Y, \Sc}{\ratio{sep}}$ and $\Ex{\Y, \Sc}{\ratio{suf}}$,
\begin{align}
  \aratio{sep} = \frac{1}{\N} \sum_{i=1}^\N
    \frac{\clf{1}{\Yi_i, \Sci_i}}{1 - \clf{1}{\Yi_i, \Sci_i}}.
    \frac{1 - \clf{1}{\Yi_i}}{\clf{1}{\Yi_i}}, \qquad
  \aratio{suf} = \frac{1}{\N} \sum_{i=1}^\N
    \frac{\clf{1}{\Yi_i, \Sci_i}}{1 - \clf{1}{\Yi_i, \Sci_i}}.
    \frac{1 - \clf{1}{\Sci_i}}{\clf{1}{\Sci_i}}.
    \label{eq:a_sep_suf}
\end{align}
These have an intuitive interpretation; the separation and sufficiency
approximations work by determining how much \emph{more} predictive power the
joint distribution of $\Y$ and $\Sc$ has in determining $\A$ over just
considering the marginals, $\Y$ or $\Sc$ respectively. Similarly for
independence, how much more predictive of $\A$ the score is, $\Sc$, over base
rate distribution $\pr{\A}$. We demonstrate this interpretation in \S\ref{sub:sim}.
Naturally this approach requires the classifiers (and data) used to be
sufficiently expressive and well calibrated\footnote{That is, we require the
use of a proper loss so it is possible to recover the posterior class
probability from the classifier. For example, log-loss, cross entropy loss,
Brier loss etc.} such that they can model the distributions $\pc{A}{\cdot}$
with little error~\cite{qin_1998, zheng_2016}. In practice we will always have
some error, and unfortunately it is difficult when using cross validation to
determine if the classifier error is because of model `miss-specification' or
because our predictor score, $\Sc$, is fair.
Working with density ratios such as these directly also has some further
limitations. For instance, it is awkward when we have categorical sensitive
attributes as we would have to look at, or aggregate, $K(K-1)/2$ density ratios
where $K$ is the number of categories. Estimating these ratios empirically also
suffers from the problem that their value can be dominated by a single instance
with vanishing probability in the denominator. Furthermore, it is also
difficult to intuit what a \emph{maximally} unfair score, or unfairness upper
bound would be if a finite bound did exist.

\section{Mutual Information Estimation}%
\label{sec:mi}

An alternative approach to measuring the fairness criteria that circumvents
some of the limitations of the direct density ratio approach is to calculate
the mutual information (MI)~\cite{gelfand_1957, cover2006} between variables.
This approach also naturally handles categorical sensitive attributes,
$\As = \{0, \ldots, K\}$.
For instance, to asses the \emph{independence} criterion from~\eqref{eq:ind},
we can calculate the MI between $\Sc$ and~$\A$,
\begin{align}
  \MI{\Sc}{\A} = \int_\Scs \sum_{\Ai \in \As} \pr{\Sci, \Ai} 
  \log \frac{\pr{\Sci, \Ai}}{\pr{\Sci}\pr{\Ai}} \intd{\Sci},
  \label{eq:mi_ind}
\end{align}
where we have omitted the random variables from the above distributions for
concision. Here we can see that when we have independence and achieve perfect
fairness, $\pr{\Sc, \A} = \pr{\Sc}\pr{\A}$, $\MI{\Sc}{\A} = 0$, otherwise MI
will be positive\footnote{This can be shown by applying Jensen's Inequality 
to~\eqref{eq:mi_ind}.}.
This measure naturally deals with
binary and categorical $\A$. Furthermore, we can \emph{normalise} MI by one of
its many known upper bounds so that it takes values in $[0, 1]$. One useful
upper bound in this context is the entropy of the sensitive attribute,
$\Ent{\A}$, that gives the normalised measure,
\begin{align}
  \NMI{\Sc}{\A} = \frac{\MI{\Sc}{\A}}{\Ent{\A}}, \qquad \text{where} \quad
  \Ent{\A} = - \sum_{\Ai \in \As} \pr{\Ai} \log \pr{\Ai}.
  \label{eq:nmi_ind}
\end{align}
The reason for using $\Ent{\A}$ as a normaliser is because MI and entropy are
related through $\MI{\Sc}{\A} = \Ent{\A} - \CEnt{\A}{\Sc}$, where
$\CEnt{\A}{\Sc}$ is conditional entropy, and is a measure of how much of the
information of the distribution of $\A$ is encoded by that of $\Sc$. When
$\NMI{\Sc}{\A} = 1$ then $\CEnt{\A}{\Sc} = 0$, so the distribution of $\Sc$
completely encodes all information about $\A$ --- which would be
\emph{maximally} unfair in the case of the independence
criterion\footnote{Since we can completely recover all information about the
sensitive attribute from the model predictions.}.
As in the case of the direct density ratio estimation, we have to 
resort to approximation to compute this measure. Firstly, $\Ent{\A}$ is simple to 
approximate empirically like previously,
\begin{align}
  \Ent{\A} \approx - \sum_{\Ai \in \As} \frac{\N_\Ai}{\N} \log \frac{\N_\Ai}{\N},
  \label{eq:ah_ind}
\end{align}
Where $\N_\Ai$ is the number of instances that have sensitive attribute $\A = \Ai$.
Then we can rewrite mutual information by first dividing by $\pr{\Sci}$,
\begin{align}
  \MI{\Sc}{\A} &= \int_\Scs \sum_{\Ai \in \As} \pr{\Sci, \Ai}
    \log \frac{\pc{\Ai}{\Sci}}{\pr{\Ai}} \intd{\Sci}
  \approx \frac{1}{\N} \sum^\N_{i=1}
    \log \frac{\clf{\Ai_i}{\Sci_i}}{\N_{a_i} / \N}.
  \label{eq:ami_ind}
\end{align}
Here we have estimated the integral--sum over $\Scs \times \As$ empirically. 
$\Ai_i$ is the sensitive class of the instance, $i$, and we have used
classifiers to estimate the conditional densities per instance as before.
Finally we can combine~\eqref{eq:ah_ind} and~\eqref{eq:ami_ind} to approximate
$\NMI{\Sc}{\A}$.

We can proceed along similar lines for the separation and sufficiency criteria,
though these require conditional mutual information with conditional entropy
normalisers,
\begin{align}
  \text{Separation:} \quad \NCMI{\Sc}{\A}{\Y} = \frac{\CMI{\Sc}{\A}{\Y}}{\CEnt{\A}{\Y}},
  \qquad
  \text{Sufficiency:} \quad \NCMI{\Y}{\A}{\Sc} =
    \frac{\CMI{\Y}{\A}{\Sc}}{\CEnt{\A}{\Sc}}.
  \label{eq:nmi_ss}
\end{align}
Here we can approximate the conditional entropy normalisers again using probabilistic
classifiers and estimating the integrals empirically,
\begin{align}
  \CEnt{\A}{\Y} =& - \int_\Ys \sum_{\Ai \in \As} \pr{\Yi, \Ai} 
    \log \pc{\Ai}{\Yi} \intd{\Yi},
  &\CEnt{\A}{\Sc} =& - \int_\Scs \sum_{\Ai \in \As} \pr{\Sci, \Ai}
    \log \pc{\Ai}{\Sci} \intd{\Sci}, \nonumber \\
  \approx& - \frac{1}{\N} \sum_{i=1}^{\N} \log \clf{\Ai_i}{\Yi_i},
  &\approx& - \frac{1}{\N} \sum_{i=1}^{\N} \log \clf{\Ai_i}{\Sci_i}.
\end{align}
These conditional entropy normalisers play a similar role as entropy in the
normalised independence criterion measure.
For separation
and sufficiency, if $\NCMI{\Sc}{\A}{\Y} = 1$ or $\NCMI{\Y}{\A}{\Sc} = 1$
respectively this means that $\CEnt{\A}{\Sc, \Y} = 0$ in both cases.
Intuitively, we can interpret this to mean that \emph{jointly} $\Y$ and $\Sc$
totally determine $\A$, which is by definition maximally unfair according to
both of these fairness criteria when $\CEnt{\A}{\Y} > 0$ and $\CEnt{\A}{\Sc} >
0$ respectively.
Again using empirical estimation and the three classifier approximations from
before we can compute condition mutual information for separation,
% {\small
\begin{align}
  \CMI{\Sc}{\A}{\Y} = \iint\limits_{\Ys~\Scs} \sum_{\Ai \in \As}
    \pr{\Yi, \Sci, \Ai} \log \frac{\pc{\Ai}{\Yi, \Sci}}{\pc{\Ai}{\Yi}}
    \intd{\Yi} \intd{\Sci}%, \nonumber \\
  \approx \frac{1}{\N} \sum_{i=1}^{\N}
    \log \frac{\clf{\Ai_i}{\Yi_i, \Sci_i}}{\clf{\Ai_i}{\Yi_i}},
\end{align}
% }
and sufficiency,
\begin{align}
  \CMI{\Y}{\A}{\Sc} = \iint\limits_{\Ys~\Scs} \sum_{\Ai \in \As}
    \pr{\Yi, \Sci, \Ai} \log \frac{\pc{\Ai}{\Yi, \Sci}}{\pc{\Ai}{\Sci}}
    \intd{\Yi} \intd{\Sci} %, \nonumber \\
  \approx \frac{1}{\N} \sum_{i=1}^{\N} 
    \log \frac{\clf{\Ai_i}{\Yi_i, \Sci_i}}{\clf{\Ai_i}{\Sci_i}}.
\end{align}
% Note that the above criteria differently combine the outputs of three distinct
% probabilistic classifiers separately predicting \clf{\Ai}{\Sci_i},
% \clf{\Ai}{\Yi_i, \Sci_i} and \clf{\Ai}{\Yi_i} and use the occurrence
% frequency to estimate the scalar quantity \pr{\Ai}.
We will refer to the approximated normalised (conditional) mutual information
criteria in~\eqref{eq:nmi_ind} and~\eqref{eq:nmi_ss} as $\ami{ind}$,
$\ami{sep}$ and $\ami{suf}$. 
% Simplifying these approximations further leads to,
% \begin{align}
%   \ami{sep} = 1 - \frac{\sum_{i=1}^\N \clf{\Ai_i}{\Yi_i, \Sci_i}}
%     {\sum_{i=1}^\N \clf{\Ai_i}{\Yi_i}}, \quad
%   \ami{suf} = 1 - \frac{\sum_{i=1}^\N \clf{\Ai_i}{\Yi_i, \Sci_i}}
%     {\sum_{i=1}^\N \clf{\Ai_i}{\Sci_i}}.
% \end{align}
%
Now we can directly see that these approximate measures are based on relative
classifier predictive probability. However, unlike the direct density ratio
measures, these more naturally operate with categorical sensitive attributes
using multiclass probabilistic classification, and they have an intuitive upper
bound for maximally unfair scores.

\section{Experiments}%
\label{sec:exp}

We now present some experiments on simulated and real data with the purpose of
demonstrating how these measures work in practice, and providing more clarity
into their functioning.

\subsection{Simulation}%
\label{sub:sim}

\begin{figure}[ht]
\centering
\begin{subfigure}[t]{0.45\linewidth}
  \centering
  \includegraphics[width=\linewidth]{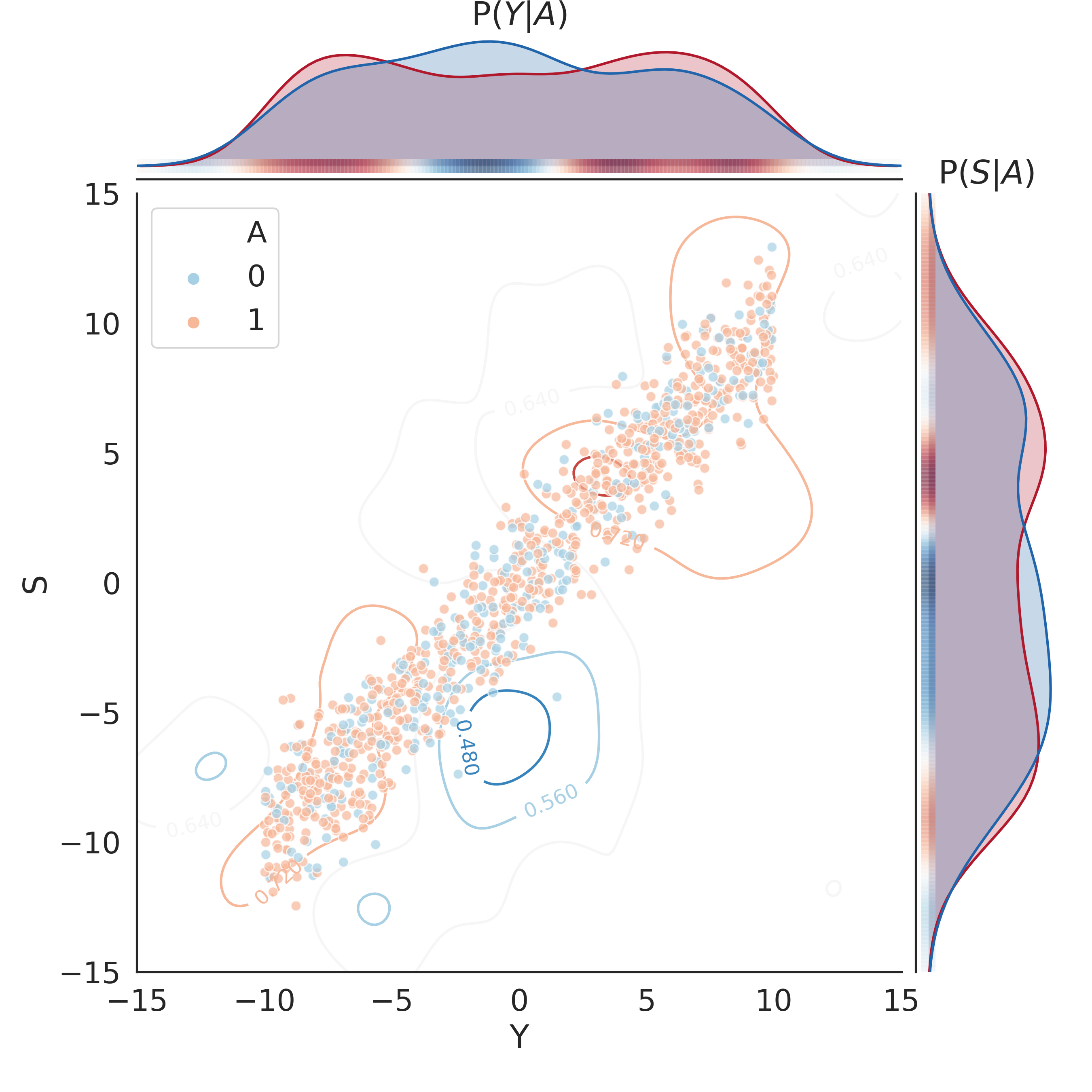}
  \caption{Fair, $\Sci_i \sim \Norm{\Yi_i}{1.5^2}$\label{fig:fair} for $\Ai_i = 0$ or $1$.}
\end{subfigure}
\begin{subfigure}[t]{0.45\linewidth}
  \centering
  \includegraphics[width=\linewidth]{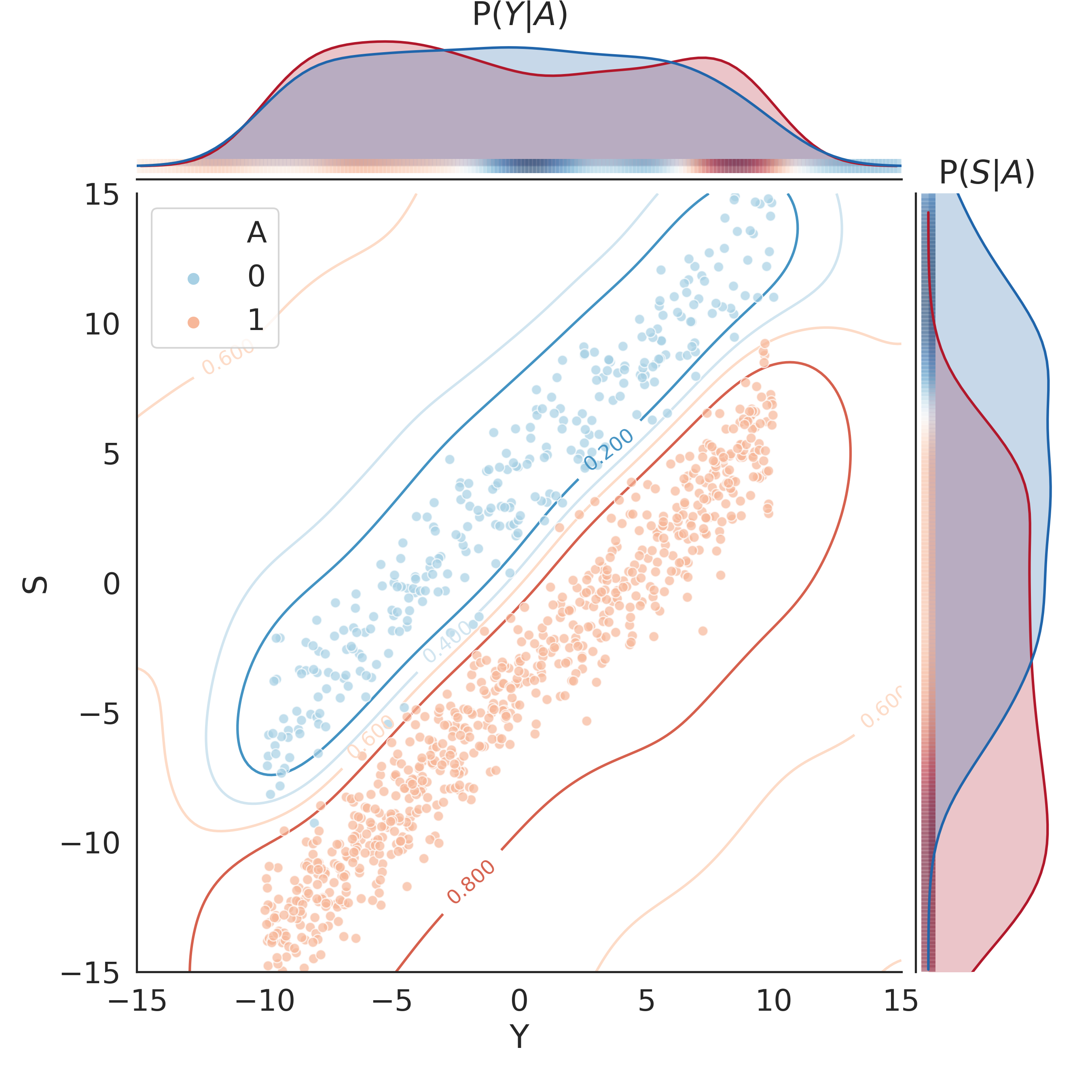}
  \caption{Score mean, $\Sci_i \sim \Norm{\Yi_i-4}{1.5^2}$ if $\Ai_i = 1$, and $\Sci_i \sim \Norm{\Yi_i+4}{1.5^2}$ if~$\Ai_i = 0$\label{fig:smean}.}
\end{subfigure}\\
\begin{subfigure}[b]{0.45\linewidth}
  \centering
  \includegraphics[width=\linewidth]{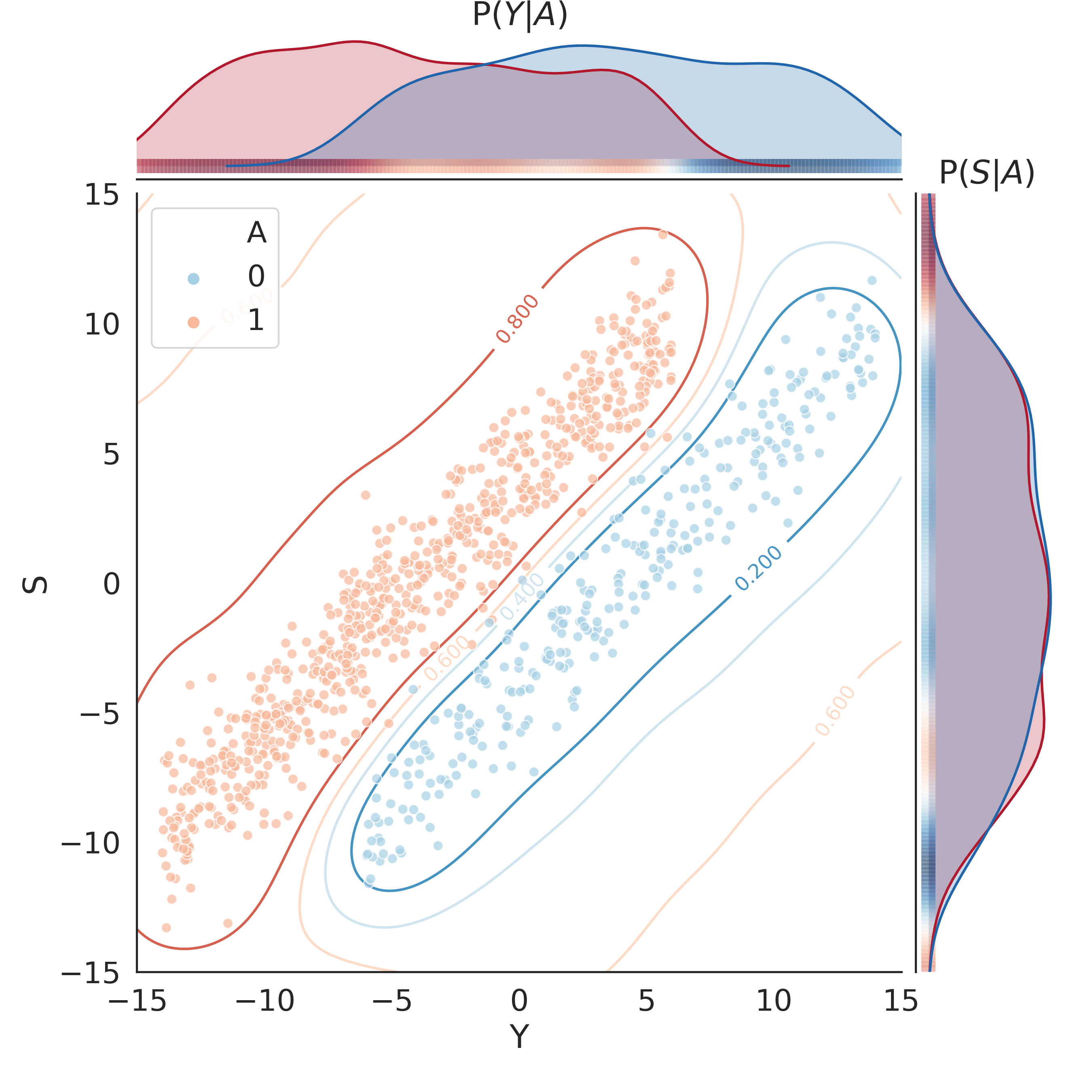}
  \caption{Target mean, $\Sci_i \sim \Norm{\Yi_i}{1.5^2}$ then $\Yi_i + 4$ \\
  if $\Ai_i = 0$, and $\Yi_i - 4$ if~$\Ai_i = 1$\label{fig:ymean}.}
\end{subfigure}
\begin{subfigure}[b]{0.45\linewidth}
  \centering
  \includegraphics[width=\linewidth]{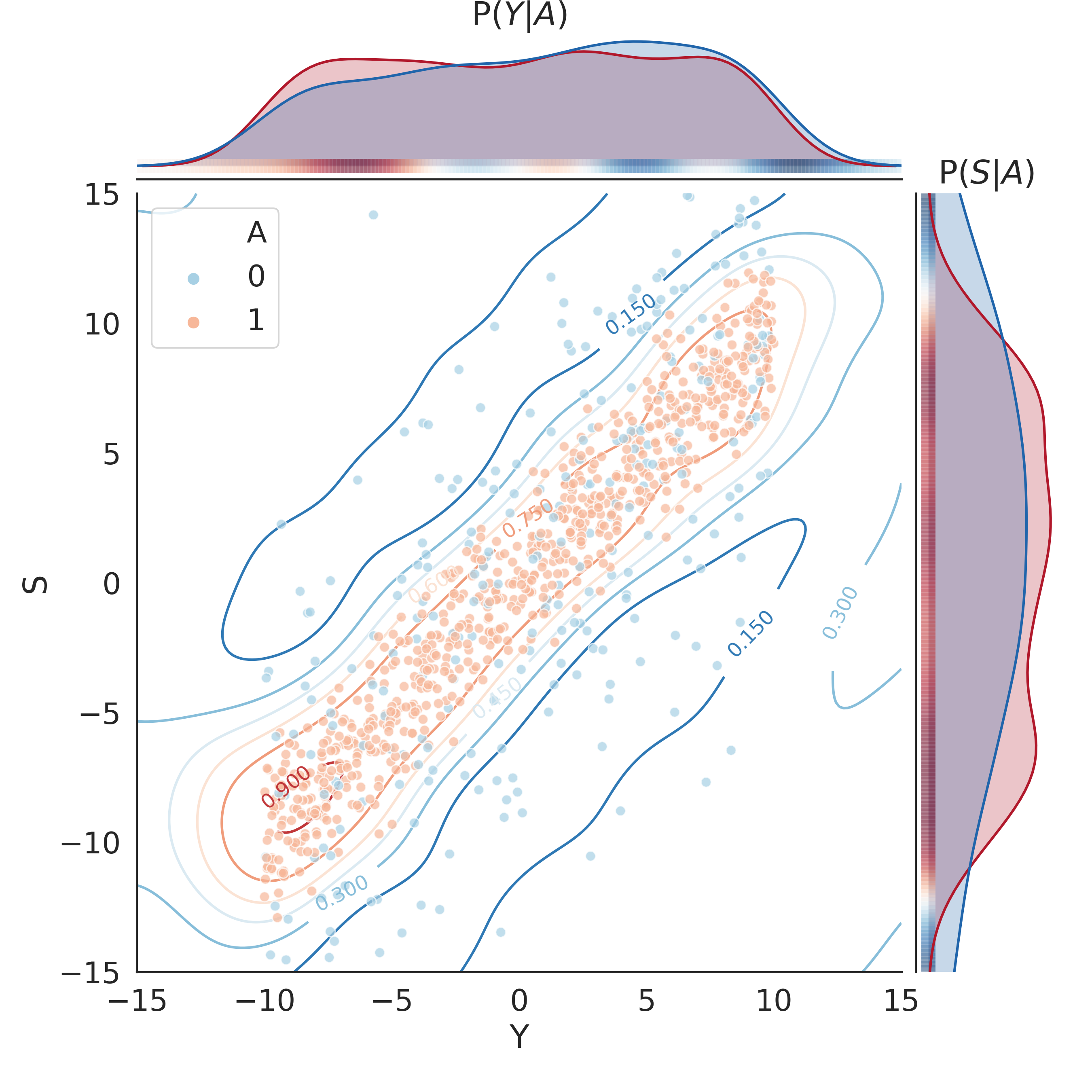}
  \caption{Score variance, $\Sci_i \sim \Norm{\Yi_i}{1.5^2}$ if $\Ai_i = 1$, and $\Sci_i \sim \Norm{\Yi_i}{6^2}$ if~$\Ai_i = 0$\label{fig:var}.}
\end{subfigure}
\caption{
  Visualisations of the four simulated data experiments demonstrating
  how the derived measures are calculated from the relative performance of 
  `competing' classifiers given different inputs.
  The $\Sci_i$ are generated according to each sub-figure caption.
  The scatter plots are the generated points in $\Ys\times\Scs$ coloured
  according to their sensitive attribute.
  We have also shown the marginal densities of these points conditioned on $\A$
  (red and blue densities).
  The contour plots show the probability contours for $\clf{1}{\Sci_i, \Yi_i}$,
  and the coloured strips in the margins represent the probabilities for
  $\clf{1}{\Yi_i}$ and $\clf{1}{\Sci_i}$.
}% TODO: more explanation
\label{fig:toy}
\end{figure}

For our first set of experiments we examine the performance of these measures
on a simulated dataset. 
The data for these experiments were generated by first drawing $\Yi_i \sim
\text{Uniform}(-10, 10)$ and $\Ai_i \sim \text{Bernoulli}(p=0.7)$ for 1000
samples. We then generated mock predictions, $\Sci_i$, by adding Gaussian
random values to $\Yi_i$ conditional on $\Ai_i$. Finally, in one experiment we
offset $\Yi_i$ conditional on $\Ai_i$. The exact data generating process for
each experiment is given in Figure~\ref{fig:toy}. We used logistic regression
with random radial basis functions to construct non-linear classifiers
$\clf{\Ai}{\cdot}$, which were validated using cross validation with 10 folds.
The results of the experiments are depicted in Figure~\ref{fig:toy}, and the
performance of the classifiers, and the values of the measures for each
experiment are in Table~\ref{tab:toy}. 

These experiments demonstrate the intuitive interpretations we gave of the
approximate measures in \S(\ref{sec:ratio}) and \S(\ref{sec:mi}).
For example, we can see from Figure~\ref{fig:ymean} that that the joint
classifier, $\clf{\Ai}{\Yi, \Sci}$ is more discriminative than the marginal,
$\clf{\Ai}{\Yi}$, which is in turn more discriminative than $\clf{\Ai}{\Sci}$.
Hence, the original predictor that produced the score, $\Sci = \fn{\x}$, is
most unfair with respect to the sufficiency criterion, then the separation
criterion.
In this instance $\clf{\Ai}{\Sci}$ is no better than random, and so
satisfies the independence criterion.

\begin{table}[tbh]
\centering
\caption{
  Classifier performance and fairness measures for the simulated experiments.
  All of these values are computed using held out classifier predictions from
  10-fold cross validation. The small negative normalised MI results are 
  numerical artefacts from the estimation process, and are not present if we
  use held-in predictive probability estimates.
}
\begin{tabular}{r|c c c|c c c|c c c}
  \toprule
  & \multicolumn{3}{c|}{Balanced Accuracy} & \multicolumn{3}{c|}{Ratio} & \multicolumn{3}{c}{Normalised MI} \\
  Experiment & \Sc & \Y & \Sc, \Y & \aratio{ind} & \aratio{sep} & \aratio{suf} &
  \ami{ind} & \ami{sep} & \ami{suf} \\
  \midrule
  Fair (\subref{fig:fair}) & .50 & 0.50 & 0.50 & 1.033 & 1.021 & 1.019 & -.003 & -.006 & -.006 \\
  Score mean (\subref{fig:smean}) & .50 & 0.65 & 0.99 & 4.489 & 9.706 & 6.912 & .271 & .890 & .847 \\
  Target mean (\subref{fig:ymean}) & .69 & 0.50 & 0.99 & 1.007 & 6.357 & 10.03 & -.015 & .841 & .898 \\
  Score variance (\subref{fig:var}) & .50 & 0.59 & 0.76 & 1.167 & 1.733 & 1.494 & .082 & .324 & .258 \\
  \bottomrule
\end{tabular}%
\label{tab:toy}%
\end{table}

\subsection{Real data} 

For validation purposes the behaviours of these measures were observed as a predictive model was tuned to reduce conditional dependencies between its predictions and a sensitive attribute. 
The Communities and Crime dataset from~\cite{Dua_2019} was used to train a linear model which included the fairness regulariser proposed in~\cite{berk_2017} in its loss function. 

The Communities and Crime dataset contains counts of all reported violent crimes for 1994 communities across the United States.
Each community instance also contains 128 demographic features from the census such as population density, average income and percentage of population that is unemployed.
For the purposes of this experiment, we have identified race as a sensitive attribute. 
Communities where more than 50\% of the population identified as Black were labelled as protected.

The group fairness regulariser in~\cite{berk_2017} penalises models whose average predictions for a group, after conditioning on the true target, depend on a sensitive attribute. 
This is a specific case of the separation criterion.

\begin{figure}[ht]
\centering
\begin{subfigure}[t]{0.45\linewidth}
  \includegraphics[width=\linewidth]{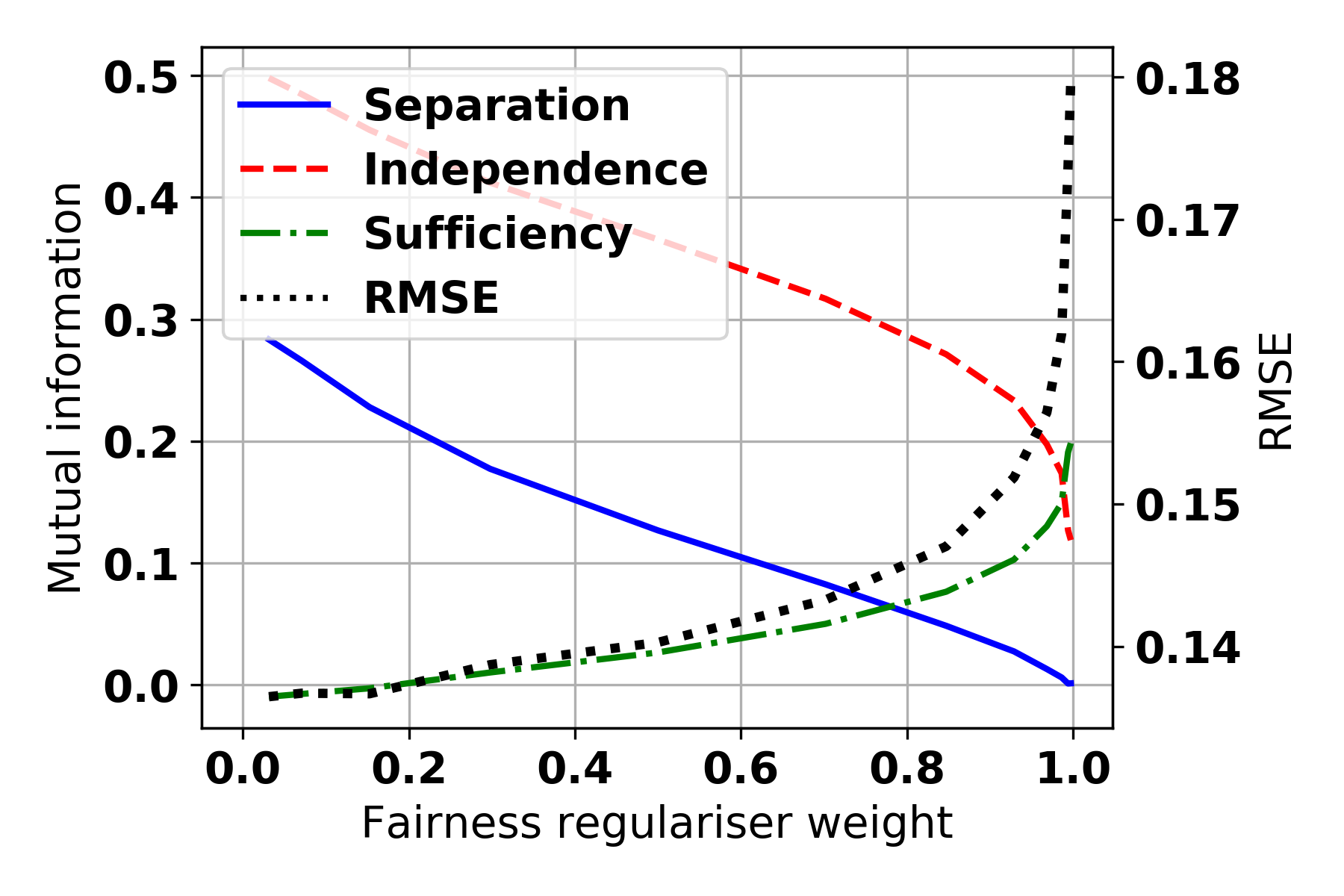}
  \caption{\label{MI_vs_reg}}
\end{subfigure}
\begin{subfigure}[t]{0.45\linewidth}
  \includegraphics[width=\linewidth]{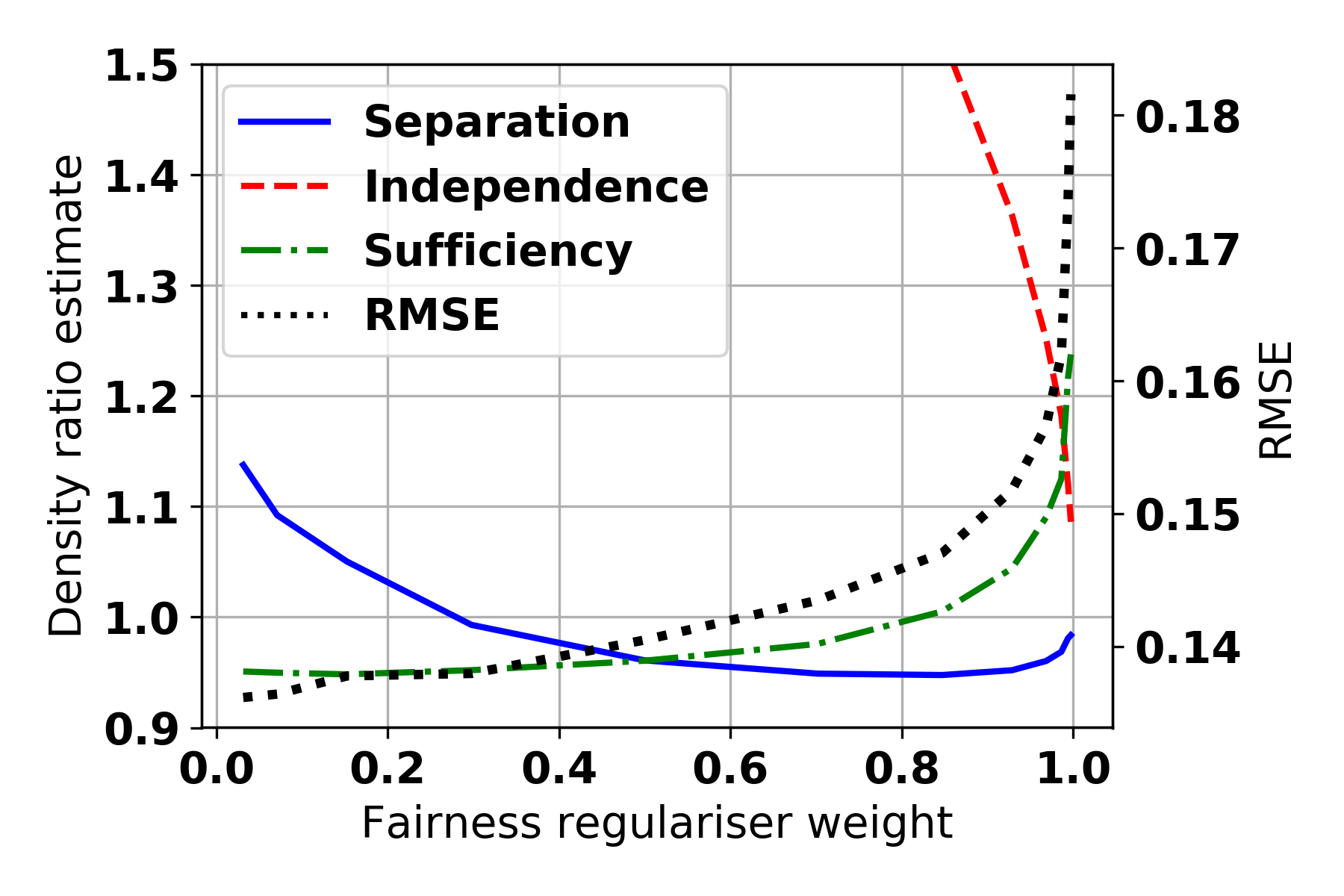}
  \caption{\label{dre_vs_reg}}
\end{subfigure} \\
\begin{subfigure}[b]{0.37\linewidth}
  \includegraphics[width=\linewidth]{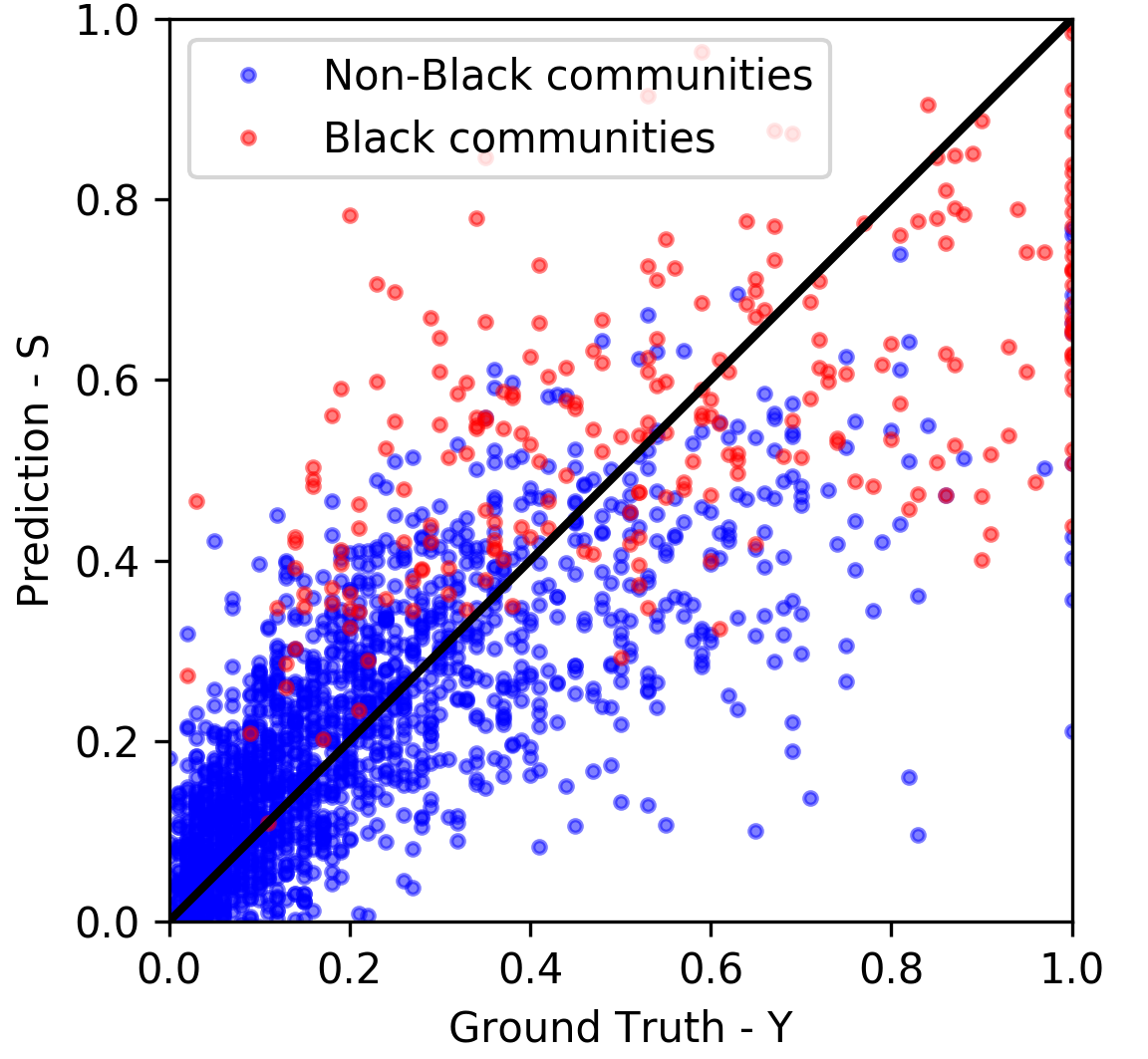}
  \caption{\label{unfair}}
\end{subfigure}
\hspace{1.3cm}
\begin{subfigure}[b]{0.37\linewidth}
  \hspace{-0.57cm}
  \includegraphics[width=\linewidth]{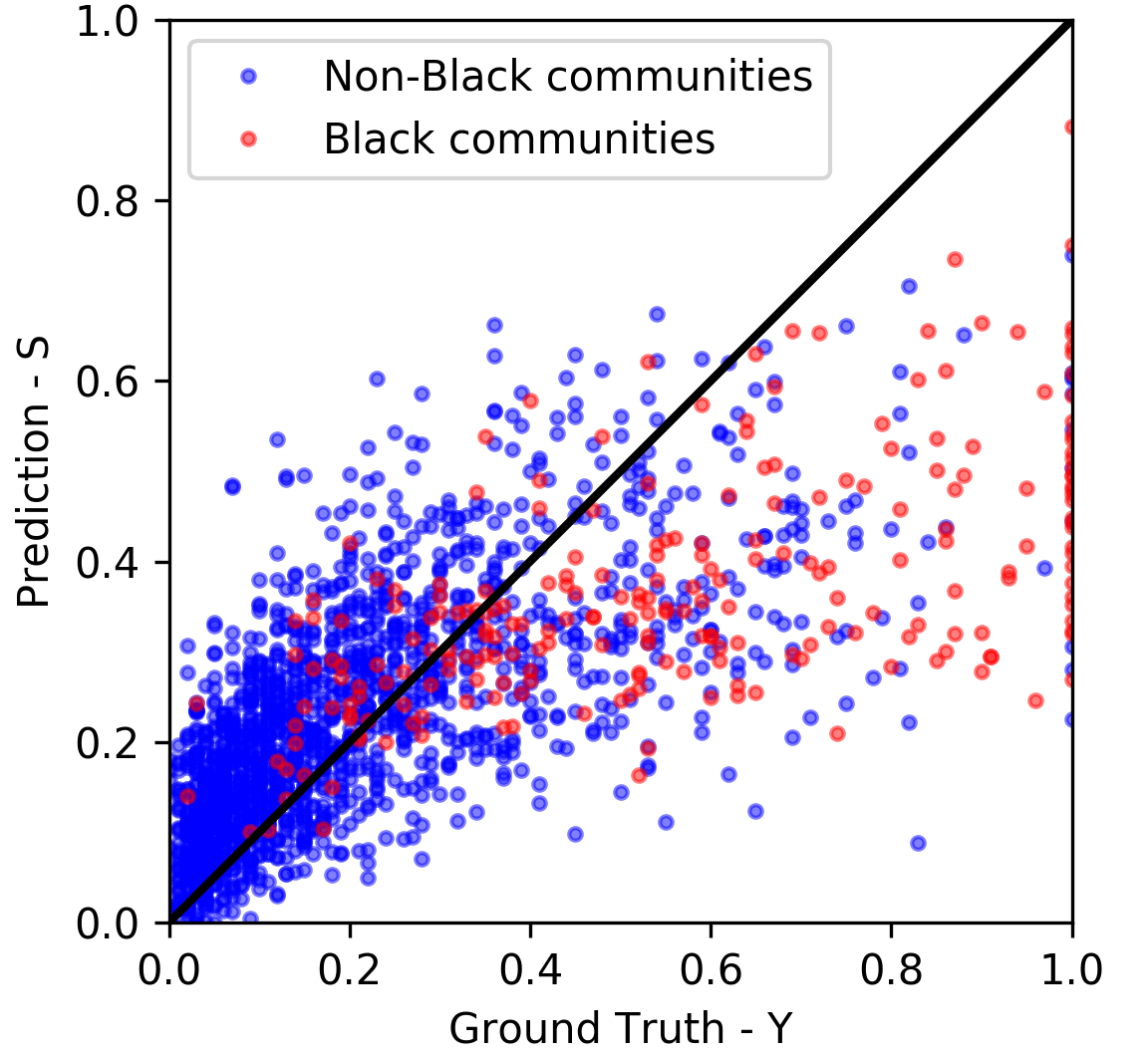}
  \caption{\label{fair}}
\end{subfigure}
\caption{\ref{MI_vs_reg} \& \ref{dre_vs_reg} Mutual information and density ratio estimates measures for the three fairness criteria  versus the normalised weight of the group fairness regulariser in \citet{berk_2017}.
\ref{unfair} $\Ys\times\Scs$-plot for a low regulariser weight. \ref{fair} $\Ys\times\Scs$-plot for a high regulariser weight.
}%
\label{fig:realdata}
\end{figure}

Figure~\ref{MI_vs_reg} \& \ref{dre_vs_reg} show the behaviour of each of the three fairness criteria for both the mutual information and density ratio estimation as the weight of the fairness regulariser is increased. 
For small regulariser weights, the loss function is dominated by its error term and consequently the
model favours maximising accuracy (minimising RMSE). 
Due to the differences in recorded crime rates in Black and non-Black communities, it is unsurprising to see that there is a strong dependence between the model's predictions and race (\emph{independence}). 
Even conditioning on the target crime rates, we still observe a dependence between the predictions and race (\emph{separation}). 
%This is an indication that the model is leveraging the difference in base rates when making its predictions.
For instance, if we condition on the targets in Figure~\ref{unfair} (a vertical
slice), we see that the model still erroneously predicts Black communities
having a higher rate of violent crimes. 
It is perhaps unsurprising that the sufficiency criterion is almost satisfied
in a model that maximises accuracy (low fairness regulariser weight). 
This is because a sufficiently unconstrained model should capture all information related to race in its predictions thus making race and the truth conditionally independent (\emph{sufficiency}).

Increasing the weight of the \citet{berk_2017}'s fairness regulariser penalises models that fail to meet the separation criteria. 
The RMSE increases due to this new constraint on the model space. As the expressiveness of the model decreases, the model can no longer fully capture information about racial base rates so conditioning on its predictions does not make race and the true targets independent. 
Thus, sufficiency is no longer satisfied --- i.e.\ the distribution of errors for a given score, now depend on race (a horizontal slice in Figure~\ref{fair}). 

Increasing the weight of the fairness regularise also improves separation as indicated by its associated mutual information and density ratio estimation measures decreasing. 
Notably, however, $\aratio{sep}$ appears to be satisfied for a lower value of the fairness regulariser than $\ami{sep}$. 
Analysis of why this is the case indicates that the approximation of the \ratio{sep} as an empirical expectation, \aratio{sep}, can lead to models that are clearly unfair, but in equal amounts, to both groups can appear fair in expectation. 
By bringing all predictions closer to the mean value (Figure~\ref{fair}), differences between groups conditioned on their true value are lessened. This also has the effect of reducing the unconditional dependence between race and predictions so the measures of independence also decreases.

\section{Discussion and future work}

The approximate density ratio and mutual information measures we derive in this
paper are simple to implement, do not depend on the algorithm used to generate
$\Sc$, and capture many properties of the conditional distributions used to
define the group fairness criteria, such as in Figure~\ref{fig:var}. They do
have limitations, as we have already noted. For instance, these approximations
are sensitive to classifier $\clf{\Ai}{\cdot}$ performance. It is also
hard to ascertain if poor classifier performance is because of a fair score, or
because of poor model choice. However, we can resort to visual inspection as in
Figure~\ref{fig:toy}, unless we have high dimensional $\Y$ and $\Sc$.
We also found the approximate mutual information measures were more
numerically robust compared to the direct density ratio measures, and were also
easier to interpret because of their normalisation.

Future directions for this work are to look at approximation methods able to
handle continuous sensitive attributes, $\As = \mathbb{R}^D$. MI approximation
methods such as LSMI~\cite{sugiyama_machine_2012} or uLSIF~\cite{kanamori_2009}
may be applicable to this task. We are also hoping that other density
ratio estimation techniques will illuminate tractable ways of incorporating
these group fairness criteria into regression objectives, or enable their use
as post processing methods for general regression algorithms.

% Tiberio, if we don't include him as an author

\small
\setlength{\bibsep}{0.7pt}
\bibliographystyle{abbrvnat}
\bibliography{references}
\end{document}